\RequirePackage{fix-cm}
\pdfoutput=1 
\documentclass[smallextended,natbib,10pt,final]{svjour3}       
\smartqed  
\usepackage{graphicx}
\AtBeginDocument{%
  \paperwidth=\dimexpr
    1in + \oddsidemargin
    + \textwidth
    + 1in + \oddsidemargin
  \relax
  \paperheight=\dimexpr
    1in + \topmargin
    + \headheight + \headsep
    + \textheight
    + 1in + \topmargin
  \relax
  \usepackage[pass]{geometry}\relax
}

\usepackage{hyperref}
\usepackage{xcolor}
\definecolor{darkblue}{rgb}{0, 0, 0.5}
\hypersetup{colorlinks=true,citecolor=darkblue, linkcolor=darkblue, urlcolor=darkblue}
\usepackage{array,longtable}

\titlerunning{WordNet-feelings}

\authorrunning{Siddharthan et al.}

\begin{document}
\date{}
\title{WordNet-feelings: A linguistic categorisation of human feelings}

\author{Advaith Siddharthan \and Nicolas Cherbuin \and Paul J. Eslinger \and Kasia Kozlowska \and Nora A. Murphy \and Leroy Lowe}

\institute{\and Advaith Siddharthan, Knowledge Media Institute, The Open University, Milton Keynes MK7 6AA, U.K. 
\email{advaith.siddharthan@open.ac.uk}\\ \and
Nicolas Cherbuin, College of Medicine Biology and Environment, 
Australian National University, 
Acton, ACT 2601, Australia. 
\email{nicolas.cherbuin@anu.edu.au}\\
\and
Paul J. Eslinger, 
Neuroscience Institute, 
Penn State Hershey Medical Center, 
Hershey, PA 17033 USA. 
\email{peslinger@pennstatehealth.psu.edu}\\
\and
Kasia Kozlowska, 
University of Sydney Medical School, 
The Children's Hospital at Westmead, 
Westmead, NSW 2006, Australia.  
\email{kkoz6421@uni.sydney.edu.au}\\
\and
Nora A. Murphy, 
Department of Psychology, Loyola Marymount University, 
Los Angeles, CA 90045, USA.  
\email{nora.murphy@lmu.edu}\\
\and
Leroy Lowe, 
President, Neuroqualia (NGO), 
Truro, Nova Scotia B2N 1X5, Canada.  
\email{leroy.lowe@neuroqualia.org}
}
\maketitle

\begin{abstract}
In this article, we present the first in depth linguistic study of human feelings. While there has been substantial research on incorporating some affective categories into linguistic analysis (e.g. sentiment, and to a lesser extent, emotion), the more diverse category of human feelings has thus far not been investigated. We surveyed the extensive interdisciplinary literature around feelings to construct a working definition of what constitutes a feeling and propose 9 broad categories of feeling. We identified potential feeling words based on their pointwise mutual information with morphological variants of the word ``feel'' in the Google n-gram corpus, and present a manual annotation exercise where 317 WordNet senses of one hundred of these words were categorised as ``not a feeling'' or as one of the 9 proposed categories of feeling. We then proceded to annotate 11386 WordNet senses of all these words to create WordNet-feelings, a new affective dataset that identifies 3664 word senses as feelings, and associates each of these with one of the 9 categories of feeling. WordNet-feelings can be used in conjunction with other datasets such as SentiWordNet that annotate word senses with complementary affective properties such as valence and intensity.
\end{abstract}

\section{Introduction}

Rosalind Picard's seminal work on Affective Computing~\cite{picard1997affective} spawned a surge in interest in topics related to feelings, emotions and affect within the computer science community. The goal has been to create intelligent systems that can simulate and recognise human-like feelings and emotions, and that has resulted in an interdisciplinary undertaking involving artificial intelligence, computational linguistics, psychology, neuroscience, and many other disciplines. Consequently, the field is becoming increasingly complex~\cite{poria2017review}, but important and fundamental conceptual hurdles remain. For example, there is still no real consensus on basic definitions for the terms ``feelings'' and ``emotions'' and although many models of emotion have been proposed, broad agreement on a comprehensive conceptual framework has been elusive~\cite{armony2013cambridge}. This lack of consistency in terminology and foundational constructs is particularly important for language processing because it leads to misunderstandings and confusion amongst researchers involved in all aspects of text analysis~\cite{munezero2014they,hovy2015sentiment,alm2012role}.

Given what we do know, some linguistics researchers have attempted to clarify the distinctions that can be made between terms such as affect, feeling or emotion~\cite{munezero2014they}. In the field of sentiment analysis, research has focused mainly on affect which has been accomplished by assigning ratings to words using basic affective dimensions such as valence (positive or negative), arousal (the level of intensity), and dominance (the degree of control exerted)~\cite{benamara2017evaluative,liu2012sentiment}. These ratings can now be found in commonly used datasets such as Affective Norms for English Words~\cite{bradley1999affective}, SentiWords~\cite{gatti2016sentiwords}, SentiWordNet~\cite{baccianella2010sentiWordNet}, and others in English~\cite{warriner2013norms,devitt2013there}, along with similar datasets in other languages~\cite{stadthagen2017norms,fairfield2017affective,monnier2017affective}.

Similarly, although there is currently no agreement on what constitutes a core set of emotions~\cite{armony2013cambridge}, and even a standing disagreement on whether or not the emotional labels we use are valid as research constructs~\cite{barrett2017theory,adolphs2017should,celeghin2017basic}, some attempts have been made to incorporate emotions into language analysis. For example, WordNet-affect 1.0~\cite{strapparava2004WordNet} is a lexical resource (based on Princeton's WordNet~\cite{miller1990introduction}) which starts with synsets that are believed to have affective content and then adds additional information about these, for example, whether they pertain to `emotion', `mood', `trait', `cognitive state', `physical state', `hedonic signal', `emotion-eliciting situation', `emotional response', `behaviour', `attitude' or `sensation'. EmoSenticNet, DepecheMood, and Topic based DepecheMood are English emotion lexicons that focus on the six emotions of `anger', `disgust', `fear', `joy', `sadness' and `surprise'~\cite{tabak2016comparison}. The Word-Emotion Association Lexicon (a.k.a. NRC Emotion Lexicon) contains lists of associations for approximately 25,000 English word senses using eight emotions (i.e., anger, disgust, fear, joy, sadness, surprise, anticipation, and trust) with automated translations~\cite{mohammad2013crowdsourcing}. Similar approaches in other languages have resulted in comparable non-English lexicons as well~\cite{stadthagen2017norms,sokolova2009classification,abdaoui2017feel}.

However, very little research exists in linguistic analysis focused on feelings (i.e., as a discrete category of language), despite the fact that feelings are gaining increasing attention in neuroscience research~\cite{damasio2013nature}. In sentiment analysis, words that convey feelings are subsumed within more generalized sets of words that are rated using affective dimensions such as valence, arousal, and dominance. Similarly, in lexicons focused on emotions, a subset of feeling words are subsumed within larger sets of words that are deemed to have affective relevance and then an attempt is made to associate them with one or more of the basic emotions being referenced (as described above). But feelings are diverse in nature and they are a fundamental part of conscious human experience~\cite{ledoux2017higher}, so the language we use to articulate them deserves careful consideration.

Confusion arises over the fact that some feelings are a component/constituent of emotional responses. For example, fear as an emotion consists of a continuum of automatically activated defense behaviors~\cite{kozlowska2015fear} that co-occur along with ``feelings of fear''. Consequently, the term feeling is often used incorrectly as a synonym for emotion and vice versa~\cite{munezero2014they,ledoux2015feelings}. But feelings are not emotions per se (which tend to be more complex~\cite{fontaine2007world}), and feelings are not limited to those that co-occur with specific emotions. Rather, feelings encompass a wide range of important mental experiences such as signifying physiological need (e.g., hunger), tissue injury (e.g., pain), optimal function (e.g., well-being), the dynamics of social interactions (e.g., gratitude), etc.~\cite{damasio2013nature,gilam2016love}.

Additional challenges relate to the fact that feelings are not consistently defined, and that our definitions for these terms can evolve over time~\cite{tissari2017current}. Moreover, while some feelings may be universally experienced across cultures (e.g., hunger, pain, cold, fatigue, etc.), other feelings are understood to be culturally constructed (e.g., gratitude~\cite{boiger2012construction}, optimism~\cite{joshi2013unrealistic}). As a result, any attempt to create a linguistic inventory of articulated feelings would need to first define feelings in a manner that can help us understand the full range of terms to be considered and then undertaken with an acute awareness that variations in terminology are going to exist in day-to-day usage, between languages, and across cultures.

In this article our goals are two-fold. At a theoretical level, we wish to compile the extensive interdisciplinary literature around feelings into  a working definition for what constitutes a feeling and to construct and define a broad categorisation of feelings that is reliable (in the sense that independent human annotators can come to similar decisions based on our definitions). Although the literature in this area is diffuse and challenging, this project was developed under the umbrella of ``The Human Affectome Project'', an initiative that began in 2016 involving a taskforce of more than 200 researchers (mainly neuroscientists and psychologists) from 21 countries. We therefore had the benefit of being able to draw upon inputs from a large pool of experts on this topic for this task. 

From a practical perspective, we also wish to create a categorised inventory of feelings. We do this by (a) identifying words in a large corpus that have a positive pointwise mutual information (PMI)~\cite{church1990word} with morphological variations of the word ``feel'', and (b) manually annotating WordNet~\cite{miller1990introduction} senses of these words with our categories of feeling.

The contributions of this article include the definitions, the categorisation, the experimental demonstration that the proposed distinctions between categories can be made by annotators, and a new dataset of WordNet synsets categorised by feeling. This new resource, WordNet-feelings, can be used in conjunction with other datasets  that annotate WordNet senses with complementary affective properties, such as SentiWordNet and WordNet-affect.

To achieve these goals we engaged a large number of researchers from the Human Affectome Project (over one hundred) both to clarify the definition of feelings and to annotate WordNet synsets. This is rather unusual, but was needed for the  validity of our definitions and the robustness of our annotations. Our approach was not just aimed at producing consensus among the six authors of this article, but also at representing the diverse interdisciplinary views outside this group bringing different theoretical perspectives and expertise.

\section{A Definition for Feelings}

To better assess the full scope of articulated feelings that would need to be included in an inventory of this nature, a definition for feelings was developed with assistance of The Human Affectome Project taskforce, as described above.  We wished to develop a comprehensive and robust functional model that could serve as a common focal point for research in the field. As such, a small task team (i.e., the authors of this article) reviewed the literature to create a definition for feelings that could serve as a starting point. We produced a first draft and shared it with the entire taskforce, feedback and input was gathered, and then the definition was refined, redistributed and the process iterated several times to achieve broad consensus within the group. The resulting definition is as follows:

\begin{quote}

A ``feeling'' is a fundamental construct in the behavioral and neurobiological sciences encompassing a wide range of mental processes and individual experiences, many of which relate to homeostatic aspects of survival and life regulation~\cite{damasio2013nature,ledoux2012rethinking,panksepp2010affective,buck1985prime,strigo2016interoception}. A broad definition for feeling is a perception/appraisal or mental representation that emerges from physiological/bodily states~\cite{damasio2013nature,ledoux2012rethinking,nummenmaa2014bodily}, processes inside (e.g., psychological processes) and outside the central nervous system, and/or environmental circumstances. However, the full range of feelings is diverse as they can emerge from emotions~\cite{damasio2013nature,panksepp2010affective,buck1985prime}, levels of arousal, actions~\cite{bernroider2011mirrors,gardiner2015integration}, hedonics (pleasure and pain)~\cite{damasio2013nature,ledoux2012rethinking,panksepp2010affective,buck1985prime}, drives~\cite{picard1997affective,alcaro2011seeking}, and cognitions (including perceptions/appraisals of self~\cite{ellemers2012group,frewen2012neuroimaging,northoff2009differential}, motives~\cite{higgins2008motives}, social interactions~\cite{damasio2013nature,ledoux2012rethinking,panksepp2010affective}, and both reflective~\cite{holland2010emotion} and anticipatory perspectives~\cite{buck1985prime,miloyan2015feelings}).

The duration of feelings can vary considerably. They are often represented in language~\cite{kircanski2012feelings} (although they can sometimes be difficult to recognize and verbalize) and some feelings can be influenced/shaped by culture~\cite{immordino2014correlations}. Feelings that are adaptive in nature~\cite{strigo2016interoception,izard2007basic} serve as a response to help an individual interpret, detect changes in, and make sense of their circumstances at any given point in time. This includes homeostatic feelings that influence other physiological/body states, other mental states, emotions, motives, actions and behaviors in support of adaptation and well-being~\cite{damasio2013nature,strigo2016interoception}. However, some feelings can be maladaptive in nature and may actually compete and/or interfere with goal-directed behavior.

A ``feeling'' is not a synonym for the term ``emotion''. There is standing debate between researchers who posit that discrete emotion categories correspond to distinct brain regions~\cite{izard2010many} and those who argue that discrete emotion categories are constructed of generalized brain networks that are not specific to those categories~\cite{lindquist2012brain}. However, both groups acknowledge that in many instances feelings are a discernable component/constituent of an emotional response (which tends to be more complex).

\end{quote}

\section{Categorising feelings}

The literature summarised above proposes several categories of feeling related to constructs of interest to different researchers, for instance: 

\begin{quote}
Physiological, 
Disgust, 
Surprise, 
Self-perceptions, 
Social (how we treat others), 
Social (how others treat us), 
Action-related, 
Proprioceptive, 
Anticipatory, 
Well-being, 
Happiness, 
Sadness, 
Fear, 
Anxiety, 
Anger, 
Arousal, 
Attention, 
Pleasure, 
Pain, 
Motivation (approach), 
Motivation (avoidance), 
Thrill/fun seeking, 
Direction of thought (anticipatory), 
Direction of thought (reflective), 
Contempt, 
Panic (flight/fight), 
Consciousness.
\end{quote}

Note that some of these categories are more specific than others, that they are biased towards the study of emotions, and that these categories are not mutually exclusive. We used our definition and these categories as a starting point to compile an initial categorisation of feelings (shown in Table \ref{tab:origcats}) that respect the distinctions proposed in the literature and are mutually exclusive. We then refined this categorisation using a data-driven approach described next.

\begin{table}[h]
\caption{Initial list of categories derived from literature}
\label{tab:origcats}
\begin{tabular}{p{0.95\textwidth}}
{\small
 Physiological/Bodily states,
 Actions,
 Anticipatory,     
 Arousal,
 Social,
 Hedonics (pleasure),
 Hedonics (pain),
 Motivivation (approach),
 Motivation (neutral),
 Motivivation (avoidance),
 General Well-Being (positive),
 General Well-Being (negative),
 Self,
 Other.}\\
 \hline
\end{tabular}

\end{table}


\subsection{Method}

While the categories in Table \ref{tab:origcats} are derived from and aim to remain faithful to distinctions made in neurological and psychological literatures, our approach to categorising feelings was additionally based on an analysis of linguistic data and an empirical assessment of the ability of human annotators to categorise such data. We began by identifying a set of ``potential feeling words'', i.e. a set of words that would together provide good coverage of the word senses that fit our working definition of a feeling. We obtained this set using the English 5-grams from the Google Books Ngram Corpus Version 2, compiled from over 4.5 million English books containing close to half a trillion words~\cite{lin2012syntactic}. We calculated for each word $x$ in this dataset, its pointwise mutual information with morphological variants of ``feel'' (feel, feeling, feelings, feels, felt), using the formula:
\[pmi(x,feel)=\frac{p(x,feel)}{p(x)p(feel)},\]
where the probabilities are obtained through maximum likelihood estimation as follows:
\begin{itemize}
\item $p(x)$ is the fraction of 5-grams containing $x$
\item $p(feel)$ is the fraction of 5-grams containing any of the above variants of ``feel''
\item $p(x,feel)$ is the fraction of 5-grams containing both $x$ and a variant of ``feel''
\end{itemize}

We collected all words $x$ for which $pmi(x,feel)>0$, i.e., all words that occur in the same 5-gram as a variant of ``feel'' more often than we would expect if their occurrences were independent.\footnote{This is a rather weak threshold, chosen to achieve wide coverage of the range of feelings, and we expect large numbers of spurious words in this list.} We then applied morphological analysis using the Xerox PARC tools~\cite{beesley2003finite}\footnote{\url{http://open.xerox.com/Services/fst-nlp-tools/Pages/morphology}} to group inflectional variants according to lemma, to obtain a file with entries such as:
\begin{quote}
\textbf{dread}: dread/+Adj; dreadful/+Adj; dreadfully/+Adv; dreading/+Adj; dreads/+Verb+Pres+3sg
\end{quote}

We then identified the WordNet~\cite{miller1990introduction} senses pertaining to each inflectional variant. The above variants of ``dread'' feature in seven synsets. Some examples of senses include:

\begin{enumerate}
\item 
\texttt{WID-00193799-A-??-dreadful}\\
\textit{dread (adjective)} (awful, dire, direful, dread, dreaded, dreadful, fearful, fearsome, frightening, horrendous, horrific, terrible) \textbf{causing fear or dread or terror;} ``the awful war''; ``an awful risk''; ``dire news''; ``a career or vengeance so direful that London was shocked''; ``the dread presence of the headmaster''; ``polio is no longer the dreaded disease it once was''; ``a dreadful storm''; ``a fearful howling''; ``horrendous explosions shook the city''; ``a terrible curse''; 

\item
\texttt{WID-01803247-A-??-dreadful}\\
\textit{dreadful (adjective) }(dreadful) \textbf{very unpleasant;} 

\item
\texttt{WID-00056340-R-??-dreadfully}\\
\textit{dreadfully (adverb)} (dreadfully, awfully, horribly) \textbf{of a dreadful kind;} "there was a dreadfully bloody accident on the road this morning"; 

\item
\texttt{WID-01780202-V-??-dread}\\
\textit{dread (verb)} (fear, dread) \textbf{be afraid or scared of; be frightened of;} "I fear the winters in Moscow"; "We should not fear the Communists!"; 
\end{enumerate}

We then iteratively (re) defined our categories and their definitions, as well as guidelines for performing the task through the following process: 
\begin{enumerate}
\item two or more humans independently annotated all the senses of 20 randomly selected words from the list with one of the feeling categories, or ``not-a-feeling'' using a web interface specially created for the purpose. 
\item we investigated inter-annotator disagreements and proposed modifications to our categories and definitions, and the task guidelines, aimed at resolving these.
\end{enumerate}

We continued this process until we were confident that remaining disagreements could not be resolved through further adaptation of the categories and their definitions. This might occur for example because of insufficient detail in a WordNet definition, different mental conceptualisations of the categories by different annotators, or just annotator error.

Summarising the changes from the set of categories we began with (Table \ref{tab:origcats}), we found that for some word senses (e.g. relating to \textit{excitement}), it was difficult to distinguish between `arousal', `anticipation' and `hedonics'. This led to merging `arousal' with `actions' to create a category `actions and prospects' and a revised definition of the term arousal within the context of actions. We also decided not to make distinctions based on valence (as such information is available from other resources such as SentiWordNet), thus merging `pain' and `pleasure' into a single `hedonics' category, and `approach' and `avoidance' into a single `attraction and repulsion' category. We then created  new categories for `anger' and `attention', as these did not fit well within our existing categories, as evidenced by recurring disagreements during annotation. Finally, we merged the `self' and `other' categories as it was difficult to enumerate all aspects of the self that feelings could pertain to and, also, the latter category was rarely used. This process produced the final set of categories shown in Table \ref{tab:cats} along with their definitions.
They aim to respect the distinctions proposed in the literature, organising them into a smaller set of distinctions that can be reliably made and are mutually exclusive. 

Our finalised task guidelines were:

\begin{enumerate}
\item Please identify whether each sense of the word is a feeling, and if so its category.
\item 
Note that verbs are presented in the present tense, but the feeling is often better expressed by the past tense and you are encouraged to think of the past tense for all verbs when deciding.
\item 
Also note that for any listed sense, the definition and accompanying examples can pertain to physical objects or other people. You need to decide if that sense of the word is nonetheless a feeling when it pertains to the self.
\end{enumerate}

\begin{figure}[t]
\includegraphics[width=\textwidth]{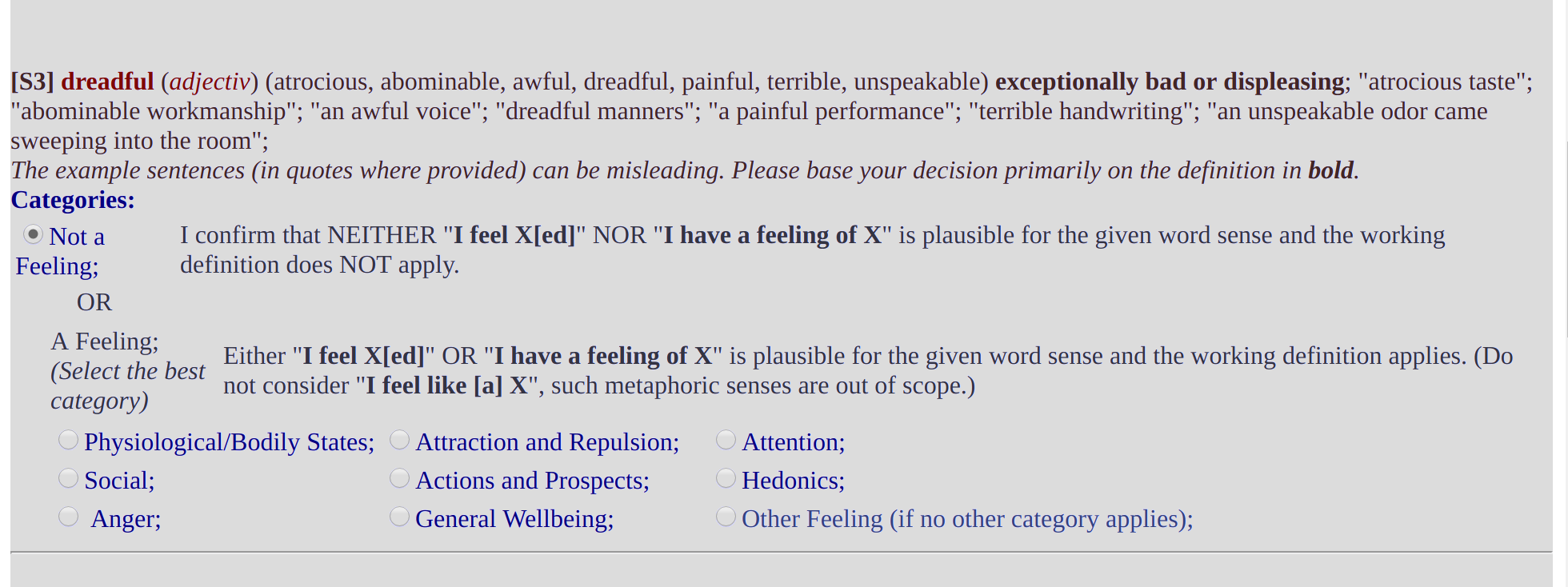}
\caption{Screenshot of annotation interface}
\label{fig:screenshot}
\end{figure}

The annotation interface also reminded annotators about key aspects of performing the task. An example is shown in the screenshot in Figure \ref{fig:screenshot}, which reminds annotators to base their judgement on the definition rather than the examples, to explicitly ask themselves whether the constructions ``I feel X[ed]'' or ``I have a feeling of X'' are plausible, and to exclude metaphoric constructs such as ``I feel like [a] X''.

{
\renewcommand{\arraystretch}{1.3}
\begin{longtable}{>{\raggedright}p{2.5cm}p{8.2cm}}
\caption[Ten categories of feeling]{The ten categories used in the study and their definitions}\label{tab:cats}\\

\textbf{Category}&\textbf{Scope}\\
\hline
\endfirsthead
\multicolumn{2}{r@{}}{Table \ref{tab:cats} continued \ldots}\\
\textbf{Category}&\textbf{Scope}\\
\hline
\endhead
\multicolumn{2}{r@{}}{continued \ldots}\\
\endfoot
\hline
\endlastfoot
Physiological or Bodily states (\texttt{Physio})&
Feelings related to specific physiological/bodily states (e.g.hungry,warm,nauseus) include feelings that relate to the current status of mental function (e.g.dizzy, forgetful, etc.)
and feelings related to energy levels (e.g. vital, tired).
However this category does not include levels of arousal (e.g., excited, relaxed, etc.)
\\
Attraction and Repulsion (\texttt{Attract})&
Feelings of attraction (e.g. love, attracted, hooked, etc.)
or repulsion (e.g. dislike, disgusted, etc)
\\
Attention (\texttt{Attent})&
Feelings related to focus, attention or interest (e.g. interested, curious, etc), or the lack of focus, attention or interest (e.g. uninterested, apathetic, etc)
\\Social (\texttt{Social})&
Feelings related to the way a person interacts with others (e.g. accepting, ungrateful, etc.). feelings related to the way others interact with that person (e.g. appreciated, exploited, trusted, etc.), or feelings of one person for or towards others (e.g. sympathy, pity, etc.) that are not covered by other categories (specifically, does not include feelings of Anger, Fear, Attraction or Repulsion).
\\Actions and Prospects (\texttt{Action})&
Feelings related to goals, tasks and actions (e.g. purpose, inspired), including feelings related to planning of actions or goals (e.g., ambitious), feelings related to readiness and capacity of planned actions (e.g. ready, daunted), feelings related to levels of arousal, typically involving changes to heart rate, blood pressure, alertness, etc.,  physical and mental states of calmness and excitement (e.g. relaxed, excited, etc.), feelings related to a person's approach, progress or unfolding circumstances as it relates to tasks/goals within the context of the surrounding environment (e.g. organized, overwhelmed, surprised, cautious, etc.), feelings related to prospects (e.g. afraid, anxious, hopeful, tense, etc.).

\ \ \ \ This category does not include feelings pertaining to Attention, (e.g. curious), Physiological energy levels (e.g. refreshed), or Social feelings that reflect attitudes towards others.

\\Hedonics (\texttt{Hedon})&
Feelings that relate to pleasurable and painful sensations and states of mind, where pleasurable includes milder feelings related to comfort and pleasure (e.g. comfortable, soothed, etc.) and painful likewise includes feelings related to discomfort and suffering (e.g.suffering, uncomfortable, etc.) in addition to pain.

\ \ \ \ This category does not include feelings of Anger, Fear, Attraction, Repulsion or General Wellbeing
\\Anger (\texttt{Anger})&
All forms of anger, directed towards self, others or objects / events (e.g. rage, anger, etc).
\\General Well-Being (\texttt{Well})&
Feelings that relate to whether or not someone is happy, content, or sad.
Feelings of general wellness that refer in a non-specific way to how someone is feeling overall (e.g. great, good, okay, fine, bad, terrible, etc.). If someone used one of these general overarching terms to describe their overall wellness, further questions would be needed to uncover the underlying (more specific) feelings that are contributing to their overall assessment of their general wellness.

\ \ \ \ This category is only for ``general'' terms and should not be used when a more specific category applies.
\\Other (\texttt{Other})&
If none of the above categories apply, but nonetheless,
the sentence ``I feel X[ed]'' is plausible for the given word sense. This category includes feelings related to appraisals of the self with respect to categories such as:
size (e.g. big, etc.), weight (e.g. fat, etc.), 
age (e.g. old, etc.), gender (e.g. masculine, etc.), 
fitness (e.g. unfit, etc.), intelligence (e.g. smart, etc.),
attractiveness (e.g. beautiful, etc.), dress and adornment (e.g. fashionable, etc.)
uniqueness (e.g. unremarkable, etc.), general normality (e.g. weird, etc.)
self-esteem (e.g. self-loathing, etc.)
identity and belonging (e.g. Buddhist, American)
\\Not a feeling (\texttt{Not})&
This category is only to be used when the working definition of a feeling does not apply to this word sense, neither ``I feel X[ed]'' nor ``I have a feeling of X'' is plausible for the given word sense, and none of the above categories fit either.

\ \ \ \ Note that this is expected to be a common case as the words you annotate can have many different senses and not all (or indeed any) need to be feelings.\\
\hline
\end{longtable}
}
\section{Human Annotation Experiment}
\label{sec:humanannot}

In our experiment, six annotators (the authors of this article) independently annotated 100 words (with 317 senses), randomly selected from the dataset, none of which had been seen during iterations aimed at finalising categories and definitions. The annotators used the working definition and category definitions provided in this article, and the web interface which presented instructions and each sense of a word in the format shown in Figure \ref{fig:screenshot}.

\subsection{Distribution of Categories}

The distribution of categories is
very skewed, with 73\% of the sense annotations belonging to the category ``Not a feeling''. This is to be expected as (a) we had set a very low threshold for collecting feeling words ($pmi > 0$) in order to ensure good coverage of feelings, resulting in many spurious words, and (b) WordNet provides very fine grained sense distinctions and therefore even for good candidate words, several senses might not be feelings. The relative frequency of each of the feeling categories in the annotation is
listed in Table~\ref{fig:percent}. Among the categories of feeling, the {\tt Social} category was most frequent, and {\tt Anger} the least frequent. Note that these are relative frequencies of word senses in our sample of 100 words. They reflect both the range of vocabulary used to express each category of feeling and the number of WordNet senses these words have. The relative frequencies of the word senses in the corpus (indicating how commonly each category of feeling is expressed through language) is likely to be very different. Estimating this is beyond the scope of this article as it would require accurate word sense disambiguation for feeling words.

\begin{table}[h]
\caption{Distribution of Feeling Categories.}
\label{fig:percent}
\begin{small}
\begin{tabular}{|c|ccccccccc|}
\hline
&\hspace*{0mm}{\tt\small 
Action}\hspace*{-2mm}  & \hspace*{-1mm}{\tt\small 
Anger}\hspace*{-1mm}  & \hspace*{-2mm}{\tt\small 
Attent}\hspace*{-1mm} & \hspace*{-2mm}{\tt\small 
Attract}\hspace*{-1mm}&\hspace*{-2mm}{\tt\small
  Hedon}\hspace*{-1mm}&\hspace*{-1mm}{\tt\small
  Other}\hspace*{-1mm}&\hspace*{-1mm}{\tt\small
  Physio}\hspace*{-1mm}&\hspace*{-1mm}{\tt\small
  Social}\hspace*{-1mm}&\hspace*{-1mm}{\tt\small
  Well}\hspace*{-1mm}
  \\
\hline
\%&11.5&0.7&1.1&3.7&5.2&13.0&13.7&34.8&16.3\\
\hline
\end{tabular}
\end{small}
\end{table}

\vspace{-10pt}
\subsection{Inter-Annotator Agreement}
\label{sec:inter}

We report inter-annotator agreement for three categorisation tasks. Following~\citet{carletta1996assessing}, we measure agreement in Cohen's $\kappa$~\cite{cohen1960coefficient}, which
  follows the formula $\kappa = \frac{P(A) - P(E)}{1 - P(E)}$ where P(A) is
  observed agreement and P(E) expected agreement.  The range of $\kappa$ if from -1
  to 1. A value of $\kappa$=0 indicates that agreement is only as expected by chance and $\kappa$=1 indicates perfect agreement.
  
For the 10-way (Actions, Anger, Attention, Attraction, Hedonics, Not-a-Feeling, Other, Physiological, Social, Wellbeing) categorisation performed by annotators at the level of word senses, we reached an inter-annotator agreement of $\kappa$=0.494 (P(A) = 0.714; n=10; N=317; k=6).

In an attempt to determine how well our working definition of a feeling performed, we
created an artificial split\footnote{This is a reasonable split to make because  the interface (cf fig. \ref{fig:screenshot}) explicitly asked annotators to first decide if a word sense constituted a feeling, before categorising it.} of the data into a binary distinction: The Not-a-feeling 
category versus a ``Feeling'' super-category consisting of all the nine feeling categories  (Anger Other Wellbeing Actions Attention Attraction Social Hedonics Physiological).
For this binary categorisation of word senses (Feeling, Not-a-Feeling), we achieved $\kappa$=0.624 (P(A) = 0.828; n=2; N=317;k=6). 

Finally, we also conflated all senses of a word to create a binary categorisation at the word level, creating word level annotations indicating whether any sense of a word is a feeling. For this task, we achieved $\kappa$=0.687 (P(A) = 0.878; n=2; N=100;k=6). 

As expected, agreement was higher for the binary categorisation than for the finer grained 10-way categorisation, and also agreement was higher for annotations at the level of words than for finer grained sense distinctions. 

In attempting to interpret these results, we first note that there does not exist any consensus for what is an acceptable value of $\kappa$, as this statistic reflects the difficulty of the categorisation task as much as anything. A commonly used interpretation comes from~\citet{landis1977measurement}, who suggested the kappa result be interpreted as follows: values  $\leq 0$ as indicating no agreement;  0.01--0.20 as slight, 0.21--0.40 as fair, 0.41--0.60 as moderate, 0.61--0.80 as substantial, and 0.81--1.00 as almost perfect agreement.~\citet{landis1977measurement} themselves note that these benchmarks though useful are arbitrary. Among the factors that can influence the magnitude of kappa are prevalence and bias~\cite{sim2005kappa}. For the same percentage agreement:
\begin{itemize}
\item When the prevalence of one or more categories is high, chance agreement is also high and kappa is reduced accordingly. 
\item When annotators exhibit different biases (i.e. favour different categories), chance agreement is reduced and kappa is higher accordingly.
\end{itemize}

Our data were clearly skewed with respect to prevalence, with 73\% in the `not a feeling' category. We also found significant annotator bias. Table \ref{tab:chi2} shows the number of times each annotator (J1--J6) has used each category. A chi-sq test for independence confirmed that the proportion of word senses assigned to each category differs from annotator to annotator ($\chi^2 (45, 317) =181.6$; $p<0.00001$). These differences are evident from the table. J2 and J4 were more conservative than the others in assigning any of the feeling categories. J1 favoured the `Other' category more than the others and J6 the `Social', etc. 

These considerations require us to take care when we compare our results to previous studies. To make meaningful comparisons even harder, previous studies on annotating word senses with affective or sentiment labels have not reported inter-annotator agreement at all, so we cannot compare our findings to those of the most directly related works. We have however found studies on classifying sentences according to emotion.~\citet{melzi2014patient} reported a study where 150 sentences from health forums were manually categorised by 6 annotators for 6 emotions (happiness, sadness, anger, disgust, surprise and fear)~\cite{ekman1992argument}. They report inter-annotator agreement of $\kappa=0.26$, considerably lower than our results. On the other, hand annotation studies about sentiment tend to report higher agreement than us, for example,~\citet{wilson2009recognizing} report a value of $\kappa=0.72$  where 2 participants label  447 subjective expressions according to their sentiment with four contrasting labels (neutral, positive, negative, both), and~\citet{o2009topic} report $\kappa=0.71$ for categorising sentences as positive, negative or neutral.

\begin{table}[t]
\caption{Distribution of categories used by each annotator J1--J6.}
\label{tab:chi2}
\begin{tabular}{|c|cccccccccc|}
\hline
&\hspace*{0mm}{\tt\small 
Action}\hspace*{-2mm}  & \hspace*{-1mm}{\tt\small 
Anger}\hspace*{-1mm}  & \hspace*{-2mm}{\tt\small 
Attent}\hspace*{-1mm} & \hspace*{-2mm}{\tt\small 
Attract}\hspace*{-1mm}&\hspace*{-2mm}{\tt\small
  Hedon}\hspace*{-1mm}&\hspace*{-1mm}{\tt\small
  Other}\hspace*{-1mm}&\hspace*{-1mm}{\tt\small
  Physio}\hspace*{-1mm}&\hspace*{-1mm}{\tt\small
  Social}\hspace*{-1mm}&\hspace*{-1mm}{\tt\small
  Well}\hspace*{-1mm}&\hspace*{-1mm}{\tt\small
  Not}\hspace*{-1mm}
  \\
\hline

J1& 	28&	0&	1&	5&	4&	38&	18&	23&	10&	190\\
J2& 	11&	0&	0&	2&	8&	4&	7&	27&	8&	250\\
J3& 	32&	0&	1&	3&	3&	24&	9&	29&	14&	202\\
J4& 	23&	1&	2&	2&	8&	4&	5&	31&	9&	232\\
J5& 	25&	1&	5&	6&	25&	20&	24&	23&	5&	183\\
J6& 	26&	0&	7&	7&	8&	21&	18&	42&	21&	167\\

\hline
\end{tabular}
\vspace{-10pt}
\end{table}

Taking into account the number and complexity of the categories and the complex working definition of a feeling provided to annotators, and in comparision to the studies above, we consider the agreement we achieved to be relatively good. 
Still, we need to give consideration to the issues raised in this section when annotating the larger dataset.
\vspace{-10pt}

\subsection{Lessons for Dataset Construction}

As discussed above, the kappa coefficient does not itself indicate whether disagreement is due to random differences (ie., those due to chance) or systematic differences (ie., those due to a consistent pattern). \citet{reidsma2008reliability} warn that though $\kappa$ is a reliable measurement of inter-annotator agreement, {\em systematic} deviations of one or more annotators from the assumed ``truth'' can result in a skewed dataset. As shown above, our data were subject to such annotator biases. While we have been highlighting this issue here, there is no consensus on how to deal
with it. Disagreements on the difficult cases in a high-level
annotation task are unlikely to ever be purely random, because the
annotators create an internal model of the semantics of the
categories, which are bound to differ somewhat.  To minimise the
effect of such biases on the dataset and to account for the skewed prevalence of the categories, we decided to:
\begin{enumerate}
\item use a large number of annotators, 
\item solicit two annotations per word sense, 
\item adjudicate disagreements, 
\item re-annotate all cases where there was agreement on the `not a feeling' category, for which the likelihood of chance agreement is particularly high,
\item set up teams for each category to examine all the senses within their category and return those where they are unsure, 
\item check cases where synonyms belonging to the same WordNet synset are annotated differently, and 
\item independently re-annotate these returned senses and adjudicate disagreements.
\end{enumerate}

\vspace{-10pt}
\section{The WordNet-feelings Dataset}
\label{sec:dataset}

\subsection{Method}

In total, we needed to annotate 11386 senses of 4185 words organised as 3151 lemmas. Following institutional ethical approval of the study protocol, 107 participants were recruited from within a large pool of scientists associated with the Human Affectome Project. These were neuroscientists, psychiatrists and psychologists from around the world interested in human affective states. All voluntarily participated in this study without any financial compensation because they are the main beneficiaries of the study; ie., the output of this study -- the categorisation of word senses by feeling -- is of intellectual interest to them. 

For training, each participant attempted a set of 20 words (64 word senses), and then went through a spreadsheet indicating the expected categories and reasoning for these in order to align their internal models of the categories. This spreadsheet was compiled from an analysis of data generated during the earlier manual annotations by the six annotators, and was designed to include positive examples for all 10 categories.

After this training step, participants annotated as few or as many word senses as they wished. 30 participants did not proceed beyond the training phase. Of the participants who contributed to the dataset, 5 annotated fewer than 20 word senses, 13 between 21 and 100 senses, 24 between 100 and 200 senses, 23 between 200 and 500 senses, and 12 more than 500 senses. We did not store their identities, but each was assigned a unique identifier so that we could ensure the same data were not sent to the same participant repeatedly. Each word sense was  categorised by two participants independently. In cases of disagreement, a third annotator (one of two selected from among the six from the first study) adjudicated these. The identities of the original annotators were not revealed in the adjudication process (and indeed were not even recorded by the system). 

All cases where there was an agreement on the ``not a feeling'' category were re-annotated independently and disagreements adjudicated, to try to ensure that no valid feelings were missed in the annotation exercise. Note that chance agreement on this category is high, while it is negligible for all the other categories. This process led to 69 additional word senses categorised as one of the nine feeling categories.

Next, for each of the nine feeling categories, all the word senses belonging to that category were sent to a team interested in that category (recruited from the wider task force) to review, and any senses that they considered doubtful were re-annotated independently with disagreements adjudicated, as before. In total 1790 word senses were reannotated in this step, of which 976 were assigned new categories.

Finally, we inspected all annotations where different synonyms belonging to the same synset were annotated differently. Note that there are valid reasons why this might happen. for example, synset WID-00887463 (verb) with definition ``give entirely to a specific person, activity, or cause''  includes synonyms such as `give' and `devote'. In our annotations, `give' is labelled `not a feeling' due to the implausibility of constructs such as "I feel give/given/gave" while `devote' is labelled as `Attention'. 913 word senses were reannotated independently in this step and disagreements adjudicated, resulting in 393 being assigned new categories.

\vspace{-5pt}

\subsection{Characteristics of the WordNet-feelings Dataset}

Following the extensive process of annotation, adjudication, checking and re-annotation of 11386 WordNet word senses as described above,  7722 (68.2\%) were categorised as ``not a feeling''. After discarding these, we generated a new dataset ``WordNet-feelings'' that contains 3664 word senses categorised in one of 9 categories of feeling. Figure \ref{tab:distribution} provides the number of word senses in each category, and their relative proportions. It is not uncommon for different senses of a word to be annotated with different categories of feeling, indeed, this is a key motivation for annotating word senses. For example, different senses of the word `crazy' pertain to `attraction and repulsion', `physiological', `actions and prospects' and `other':

\begin{enumerate}
\item 

WID-00886448-A-??-crazy	(crazy, wild, dotty, gaga)	intensely enthusiastic about or preoccupied with; "crazy about cars and racing"; "he is potty about her"	\textbf{Attraction and Repulsion}

\item
WID-02075321-A-??-crazy	(brainsick, crazy, demented, disturbed, mad, sick, unbalanced, unhinged)	affected with madness or insanity; "a man who had gone mad"	\textbf{Physiological}	

\item
WID-01836766-A-??-crazy	(crazy, half-baked, screwball, softheaded)	foolish; totally unsound; "a crazy scheme"; "half-baked ideas"; "a screwball proposal without a prayer of working"	\textbf{Actions and Prospects}

\item
WID-00967897-A-??-crazy	(crazy)	bizarre or fantastic; "had a crazy dream"; "wore a crazy hat"	\textbf{Other}

\end{enumerate}

\begin{table}[b]
\vspace{-10pt}
\caption{Distribution of categories in WordNet-feelings}
\label{tab:distribution}
\begin{small}
\begin{tabular}{|l|ccccccccc|}
\hline
&\hspace*{0mm}{\tt\small 
Action}\hspace*{-2mm}  & \hspace*{-1mm}{\tt\small 
Anger}\hspace*{-1mm}  & \hspace*{-2mm}{\tt\small 
Attent}\hspace*{-1mm} & \hspace*{-2mm}{\tt\small 
Attract}\hspace*{-1mm}&\hspace*{-2mm}{\tt\small
  Hedon}\hspace*{-1mm}&\hspace*{-1mm}{\tt\small
  Other}\hspace*{-1mm}&\hspace*{-1mm}{\tt\small
  Physio}\hspace*{-1mm}&\hspace*{-1mm}{\tt\small
  Social}\hspace*{-1mm}&\hspace*{-1mm}{\tt\small
  Well}\hspace*{-1mm}
  \\
\hline

Number of Senses
&1160
&86
&51
&102
&108
&841
&519
&636
&161\\


\%
&31.7
&2.4
&1.4
&2.8
&3.0
&22.9
&14.1
&17.3
&4.4\\
\hline
\end{tabular}
\end{small}
\end{table}

\begin{table}[t]
\caption{Summary of WordNet-feelings in comparison to WordNet-affect}
\label{tab:comp}
\begin{tabular}{|l|cccc|}
\hline
&Adjectives&Verbs&Nouns&Adverbs\\
\hline
WN-feelings (Senses)&2385&1024&224&31\\
WN-feelings (Synsets)&1809&742&203&29\\
WN-Affect 1.0 Core (Synsets)&619&288&683&19\\
WN-Affect 1.0 All (Synsets)&1477&322&772&333\\
WN-Affect 1.1 (Synsets) &323&138&280&148\\
\hline
\end{tabular}
\vspace{-5pt}

\end{table}

WordNet-feelings is a complementary resource to other affective annotations over WordNet. It can be combined with SentiWordNet to provide additional information about valence, i.e. the degree to which the feeling is positive or negative, for all our annotations,
and with WordNet-affect, which consists of two annotations. Version 1.0 contains 2904 WordNet synsets annotated as one of `emotion', `mood', `trait', `cognitive state',
`physical state',
`hedonic signal',
`emotion-eliciting situation',
`emotional response',
`behaviour',
`attitude' or
`sensation'. These consist of a smaller set of 1609 ``core'' manual annotations, and 1295 addition synsets  automatically obtained through the use of various WordNet relations. Version 1.1 manually annotates 889 WordNet synsets with finer grained distinctions for emotions, organised hierarchically. Table \ref{tab:comp} shows the size of WordNet-feelings and WordNet-affect by part-of-speech. All datasets mainly consist of adjectives. WordNet-feelings contains a higher number of verbs and very few adverbs, and WordNet-affect a higher number of nouns. These differences can be attributed to conceptual differences between feelings and other affective categories and also to our strict guidelines for accepting a sense as a feeling only if the phrases ``I feel X[ed]'' or ``I have a feeling of X'' are plausible. Due to these guidelines, for example, senses of nouns such as ``conscience'' were labelled `not a feeling', though annotations exist in WordNet-affect.

Table \ref{tab:ex-overlap} shows some examples where there are annotations available across all datasets. The examples illustrate some differences between WordNet-feelings and WordNet-affect. Consider the first two senses in the table, for the words `disinclined' and `hostile'. WordNet-feelings categorises the first as `Actions and Prospects' as the unwillingness pertains to a persons approach, progress or unfolding circumstances and the second as `Anger' as it is hostility expressed towards others. In the third and fourth examples (`amicable' and `ardour'), WordNet-feelings distinguishes between social feelings and feelings of attraction, two categories that are the focus of much recent research in the neurosciences. In each of these cases, WordNet-affect 1.0 catgorises these at high level, such as `attitude' or `emotion', and 1.1 makes very fine-grained distinctions, which moving up the hierarchy can be interpreted as positive or negative emotions.

Table \ref{tab:ex-nooverlap} shows some examples where there are no annotations available in either WordNet-affect dataset. These span all nine categories of feelings and the table provides one example for each category.

\begin{table}[t]
\caption{Examples of annotations from WordNet-feelings alongside those from WordNet-affect and SentiWordNet}
\label{tab:ex-overlap}
{\small
\begin{tabular}{p{2.5cm}p{8.5cm}}

WID-01293158-A \textbf{disinclined}&	(disinclined)	unwilling because of mild dislike or disapproval; "disinclined to say anything to anybody"\\[3pt]
WN-feelings&Actions and Prospects\\	
SentiWN& Pos=0	Neg=0.75\\
WN-affect 1.0 &attitude	\\
WN-affect 1.1 & disinclination	[disinclination $<$ dislike $<$ general-dislike $<$ negative-emotion $<$ emotion $<$ affective-state $<$ mental-state]\\
\hline
&\\
WID-01244410-A \textbf{hostile}	& (hostile)	characterized by enmity or ill will; "a hostile nation"; "a hostile remark"; "hostile actions"	\\[3pt]

WN-feelings&Anger\\
SentiWN&Pos=0	Neg=0.625\\
WN-affect 1.0 & attitude\\	
WN-affect 1.1& hostility	[hostility $<$ hate $<$ general-dislike $<$ negative-emotion $<$ emotion $<$ affective-state $<$ mental-state]\\
\hline
&\\
WID-01246579-A \textbf{amicable}	& (amicable)	characterized by friendship and good will\\[13pt]
WN-feelings&Social	\\
SentiWN&Pos=0.875	Neg=0	\\
WN-affect 1.0&emotion-eliciting situation	\\
WN-affect 1.1&amicability	amicability $<$ friendliness $<$ liking $<$ positive-emotion $<$ emotion $<$ affective-state $<$ mental-state $<$ root
\\
\hline
&\\
WID-07544129-N \textbf{ardour}	& (ardor, ardour)	intense feeling of love\\[13pt]

WN-feelings&Attraction and Repulsion	\\
SentiWN&Pos=0.5	Neg=0.375\\
WN-affect 1.0&emotion	\\
WN-affect 1.1&love-ardor	love-ardor $<$ love $<$ positive-emotion $<$ emotion $<$ affective-state $<$ mental-state $<$ root
\\
\hline
\end{tabular}
}
\end{table}

\begin{table}[h]
\caption{Examples of annotations from WordNet-feelings where annotations are missing in both WordNet-affect 1.0 and 1.1}
\label{tab:ex-nooverlap}
{\small
\begin{tabular}{p{2.5cm}p{8.7cm}}
WID-05697789-N \textbf{certitude}	&(certitude, cocksureness, overconfidence)	total certainty or greater certainty than circumstances warrant\\[3pt]

WN-feelings&Actions and Prospects\\ 
SentiWN&Pos=0.5	Neg=0\\
\hline
&\\

WID-01788733-V \textbf{chafe}&	(chafe)	feel extreme irritation or anger; "He was chafing at her suggestion that he stay at home while she went on a vacation"	\\[3pt]
WN-feelings&Anger\\
SentiWN&Pos=0	Neg=0.5\\
\hline
&\\

WID-00600370-V \textbf{engross}	&(absorb, engross, engage, occupy)	consume all of one's attention or time; "Her interest in butterflies absorbs her completely"\\[3pt]

WN-feelings&Attention\\
SentiWN&Pos=0.125	Neg=0\\
\hline
&\\

WID-01465668-A \textbf{smitten}	&(enamored, infatuated, in\_love, potty, smitten, soft\_on, taken\_with)	marked by foolish or unreasoning fondness; "gaga over the rock group's new album"; "he was infatuated with her"	\\[3pt]

WN-feelings&Attraction and Repulsion\\
SentiWN&Pos=0.75	Neg=0\\
\hline
&\\

WID-01364585-A \textbf{tormented}&	(anguished, tormented, tortured)	experiencing intense pain especially mental pain; "an anguished conscience"; "a small tormented schoolboy"; "a tortured witness to another's humiliation"\\[3pt]

WN-feelings&Hedonics\\
SentiWN&Pos=0	Neg=0.625\\
\hline
&\\

WID-00828336-A \textbf{muscular}	&(mesomorphic, muscular)	having a robust muscular body-build characterized by predominance of structures (bone and muscle and connective tissue) developed from the embryonic mesodermal layer\\[3pt]

WN-feelings&Other\\
SentiWN&Pos=0.25	Neg=0\\
\hline
&\\

WID-01270004-A \textbf{thirsty}	&(thirsty)	feeling a need or desire to drink; "after playing hard the children were thirsty"\\[3pt]
WN-feelings&Physiological	\\
SentiWN&Pos=0.25	Neg=0.25\\
\hline
&\\

WID-01258264-A \textbf{frosty} &	(frigid, frosty, frozen, glacial, icy, wintry)	devoid of warmth and cordiality; expressive of unfriendliness or disdain; "a frigid greeting"; "got a frosty reception"; "a frozen look on their faces"; "a glacial handshake"; "icy stare"; "wintry smile"	\\[3pt]
WN-feelings&Social\\
SentiWN&Pos=0	Neg=0.875\\
\hline
&\\
WID-00363621-A \textbf{buoyant}&	(buoyant, chirpy, perky)	characterized by liveliness and lightheartedness; "buoyant spirits"; "his quick wit and chirpy humor"; "looking bright and well and chirpy"; "a perky little widow in her 70s"	\\[3pt]
WN-feelings&Wellbeing	\\
SentiWN&Pos=0.5	Neg=0.25\\
\hline
&\\

\end{tabular}
}
\end{table}
\vspace{-5pt}

\section{Conclusions}

In this article, we have described a new resource WordNet-feelings\footnote{WordNet-Feelings is available from \url{https://github.com/as36438/WordNet-feelings}}, that consists of manual annotations of 3664 WordNet senses with nine categories of feeling. To achieve this, we first had to define a feeling, a task that required us to survey the extensive interdisciplinary literature around feelings and consult a wide range of researchers. We then proposed nine categories of feeling, which respect key distinctions in the literature, are mutually exclusive, and can be used to categorise word senses reliably. We presented empirical results about the level of agreement between annotators, and proceeded to annotate a large number of WordNet senses. Throughout this process, our aim was to represent the diverse interdisciplinary view that exist both within the six authors of this article and outside of this group. Over one hundred researchers contributed towards our definition of a feeling and to the annotation of our dataset. The annotations in the data set have been made through a rigorous 
process, with independent annotations and adjudication of disagreements, as well as procedures for screening the senses in each category and re-annotating potentially problematic cases.

To our knowledge, no other research currently exists that captures this sort of an inventory of feeling words, nor is there any that attempts to define categories for such a broad range of feelings.  Although there is a close relationship between many feelings and emotions, there is currently no clear understanding of the manner in which all of these feelings are related to our many emotional responses.  So there is certainly a need for a comprehensive and robust functional model that encompasses feelings and emotions.  We recognize that  this is only one step in that direction, but we think that this initial framework should serves as a helpful starting point.
 
We do need to emphasize that this inventory of feeling words and these initial categorisations are in no way intended to be a definitive representation of the human condition.  As we noted in the introduction, linguistic variations are going to exist in day-to-day usage, between languages, and across cultures.  Nonetheless, we have much to learn in this emerging area of science, so we expect this initial dataset will be of analytical value to a wide range of researchers, including those studying feelings from a neurobiological or psychological perspective and computational linguists interested in understanding this essential part of the human condition for the purpose of text interpretation or generation.

\vspace{-10pt}

\section*{Acknowledgements}
We would like to thank all the participants in the Human Affectome Project who influenced this work through their input into the definition of a feeling and contributed their time and effort towards annotating the dataset.

\vspace{-15pt}
\bibliographystyle{spbasic}
\bibliography{refs}

\end{document}